\documentclass[10pt,twocolumn,letterpaper]{article}

\usepackage{cvpr}              

\usepackage{graphicx}
\usepackage{amsmath}
\usepackage{amssymb}
\usepackage{booktabs}

\usepackage[pagebackref,breaklinks,colorlinks]{hyperref}

\usepackage{comment}
\usepackage{multirow}
\usepackage{enumitem}

\usepackage[capitalize]{cleveref}
\crefname{section}{Sec.}{Secs.}
\Crefname{section}{Section}{Sections}
\Crefname{table}{Table}{Tables}
\crefname{table}{Tab.}{Tabs.}

\def\cvprPaperID{6476} 
\def\confName{CVPR}
\def\confYear{2022}

\begin{document}

\title{MorDeephy: Face Morphing Detection Via Fused Classification} 

\author{Iurii Medvedev$^*$\\
$^*$University of Coimbra\\
Institute of Systems \\
and Robotics\\
3030-194, Coimbra, Portugal\\
{\tt\small iurii.medvedev@isr.uc.pt}
\and
Farhad Shadmand$^*$\\
{\tt\small farhad.shadmand@isr.uc.pt}
\and
Nuno Gonçalves$^{*,\dagger}$\\
$^\dagger$Portuguese Mint and Official \\
Printing Office (INCM)\\
1000-042, Lisbon, Portugal\\
{\tt\small nunogon@deec.uc.pt}
}

\maketitle

\begin{abstract}
   Face morphing attack detection (MAD) is one of the most challenging tasks in the field of face recognition nowadays. 
   In this work, we introduce a novel deep learning strategy for a single image face morphing detection, which implies the discrimination of morphed face images along with a sophisticated face recognition task in a complex classification scheme. It is directed onto learning the deep facial features, which carry information about the authenticity of these features.
   Our work also introduces several additional contributions: the public and easy-to-use face morphing detection benchmark and the results of our wild datasets filtering strategy.
   Our method, which we call MorDeephy, achieved the state of the art performance and demonstrated a prominent ability for generalising the task of morphing detection to unseen scenarios. 
   
\end{abstract}

\section{Introduction}

Last decades with the development of deep learning techniques the evident advances have been reached in the area of face recognition. However, evolved and sophisticated techniques for performing the presentation attacks continue to appear, which require the development of new protection solutions. 

Face morphing is one such image manipulating technique. It is usually performed by blending several (usually two) digital face images and allows to match different persons with this synthetic image that contains characteristics from both faces. While being simple to implement, face morphing poses the security risks of issuing an identification document that may be validated for two or more persons. Presentation attacks with face morphing usually can be hardly detected by humans which usually perform poorly in matching unfamiliar faces on photos of ID and travel documents \cite{trustfaces_ieee_paper} and by face recognition software in ABC (automatic border control) systems \cite{magic_passport}.

In the last years, face morphing has become a matter of research interest in academia \cite{nist_conference} and industry \cite{iMARS_project}.
Morphing detection methods in facial biometric systems may be distinguished into two pipelines depending on the processing scenario. In \textit{no-reference} morphing attack detection algorithm receives a single image, where morphing is detected. In practice, these methods are directed to mitigate risks related to the false acceptance of manipulated images in the \textit{enrollment} process. The authentic document, which is generated with a successfully accepted forged image, may further help to deceive the face recognition system. 

The \textit{differential} morphing detection implies additional live data acquisition from an authentication system which gives the reference information for the detection algorithm. This scenario usually takes place while passing an Automated Border Control (ABC) system, when the recently enrolled image (which is already accepted and printed on the ID Document) is tested against morphing detection.

First morphing detection solutions relied on the behaviour of local image characteristics (like texture, noise). Recent approaches usually employ deep learning computer vision tools.
However, many of these methods utilize a straightforward learning strategy that is limited by binary classification or contrast learning, which in our opinion is not optimal for a task of face morphing detection and may lead to various convergence problems.

In this work, we introduce a novel deep learning method for single image face morphing detection, which incorporates sophisticated face recognition tasks and implies utilising a combined classification scheme (discussed in Section \ref{Methodology}).
Also, we develop the public face morphing detection benchmark, which is designed to be adaptive to the developer needs and at the same time to be simple for comparison of algorithms of different developers.
As an additional contribution, we introduce the results of our datasets filtering strategy (image name lists), which is described in Section \ref{dataset_filtering}.

Regarding the limitations of the work, it is important to note that at the current stage we focus on single image morphing detection. Also, we do not take into account redigitalized face images (by printing/scanning).
At the same time, we are limited to utilising landmark-based methods for performing face morphing. GAN (Generative adversarial Network) based methods require large computational resources (namely for projecting images to latent space) and at the same time, face recognition systems are less vulnerable to presentation attacks with GAN morphs, rather than to landmark-based morphs \cite{Can_GAN_Morphs}. However, we intend to cover those limitations in further research.

\section{Related Work}

To introduce our methodology, we need to discuss recent advances in face morphing, face morphing detection (focusing on the no-reference scenario) and face recognition. 

\subsection{Face Morphing}

The generic pipeline of creating face morph from original images includes the following steps: face features extraction $\rightarrow$ features averaging $\rightarrow$ generating morphed image from averaged features $\rightarrow$ optional restoring image context (namely background). 

Landmark based approaches, first introduced by Ferrara \etal \cite {magic_passport}, follow this pipeline straightforwardly in the image spatial domain by the face landmark alignment, image warping and blending. 
Different reported morphing algorithms employ variations of this strategy \cite{ubo_morpher,FaceFusion_morpher,learn_opencv_morpher}. 

With recent advances in generative deep learning approaches, several face morphing methods, which utilise deep latent feature domain, were proposed. 

The above face morphing pipeline may adapt various deep learning tools like variational autoencoders (VAE) \cite{morGAN} or generative adversarial networks (GANs) \cite{ Can_GAN_Morphs, MIPGAN_morphing_paper}. 

\subsection{Face Morphing Detection}

Single image (no-reference) face morphing detection algorithms usually utilize local image information and image statistics. 

Various morphing detection approaches employ Binarized Statistical Image Features (BSIF) \cite{detecting_face_morph_1}, Photo Response Non-Uniformity (PRNU), known as sensor noise \cite{PRNU_2, detecting_face_morph_PRNU}, textuxe features \cite{Towards_morphing_detection},  
local features in frequency and spatial image domain \cite{face_morphing_fd} 
or complex combination of these features \cite{morphing_fusion,face_rec_vulnerability_morph}.

Several deep learning approaches for no-reference case were proposed.
For face morphing detection these approaches usually follow binary classification of pretrained face recognition features \cite{face_morphing_dnn}, which may be finetuned \cite{detecting_face_morph__deep_learning, unibo} or utilized in a combination with local texture characteristics \cite{face_morphing_using_general_purpose_fr}.
Damer \etal \cite{PW-MAD} introduced a better regularized strategy for morphing detection by replacing the trivial binary classification with pixel-wise supervision.
Aghdaie \etal \cite{AttentionMorphing} adopted the attention mechanism which is controlled by wavelet decomposition.

Differential face morphing detection is a less challenging task and security risks in this scenario indeed may be combated by increasing the discriminability of face deep representation, which is utilized for recognition.

Several approaches for differential detection was recently proposed. Scherhag \etal \cite{face_morphing_MAD} followed the classification of pretrained deep features in differential scenario. 
Borghi \etal \cite{Double_Siamese_Morphing} performed differential morphing detection by finetuning the pretrained networks in a complex setup with identity verification and artifacts detection blocks. 
Rather different approach to the differential scenario was introduced by Ferrara \etal \cite{face_demorphing} who proposed an approach to revert morphing by retouching the testing face image with a trusted live capture to reveal the identity of the legitimate document owner.

In comparison to the considered approaches, we propose to focus our method on learning the authenticity of deep face features, regularizing the morphing detection with a delicate face recognition task.

\subsection{Face Recognition}
Modern face recognition approaches rely on deep learning tools, which give the ability to learn highly discriminative features themselves from unconstrained images. 
Among several techniques to perform the tasks of extracting features, the convolutional neural network (CNN) is one of the most efficient for the pattern recognition problems \cite{ImageNet_cite}.

There are several strategies for approaching face recognition via deep learning. However, all of them are focused on extracting low-dimensional face representation (deep face features) and increasing the discriminative power of that representation.

Metric learning methods are directed on optimising the face representation itself through the contrast of match/non-match pairs \cite{chopra_metric_paper, facenet}. 
However, for reliable convergence, these methods require enormously large datasets and sophisticated sample mining techniques. 

Another concept (which we indeed follow in our work) is learning face representation implicitly via a closed-set identity classification task. Deep networks in these methods encapsulate face representation in the last hidden layer and usually adopt softmax loss and its modifications for classification \cite{deepid_paper,deepid2_paper,deepid2_plus_paper}. 

Improvement of the performance in this technique was achieved by various techniques for increasing intra-class compactness and maximizing inter-class discrepancy. For example, by applying additional regularisation for pushing intra-class features to their centre \cite{centerface_paper}, or by introducing several kinds of marginal restrictions for inter-class variance  \cite{sphereface_paper, cosface_paper, arcface_paper, equalized_margin_paper}.

Several recent works were directed onto investigating sample specific learning strategies, which are controlled by sample quality \cite{QualFace}, hardness \cite{npcface, Huang_2020_CVPR}, data augmentation \cite{towards_face_recognition} or even by treating facial representation in distributional manner (by specifying sample \textit{uncertainty}) \cite{probabilistic_embedding}.



\begin{figure*}
\begin{center}
  \includegraphics[width=0.95\linewidth]{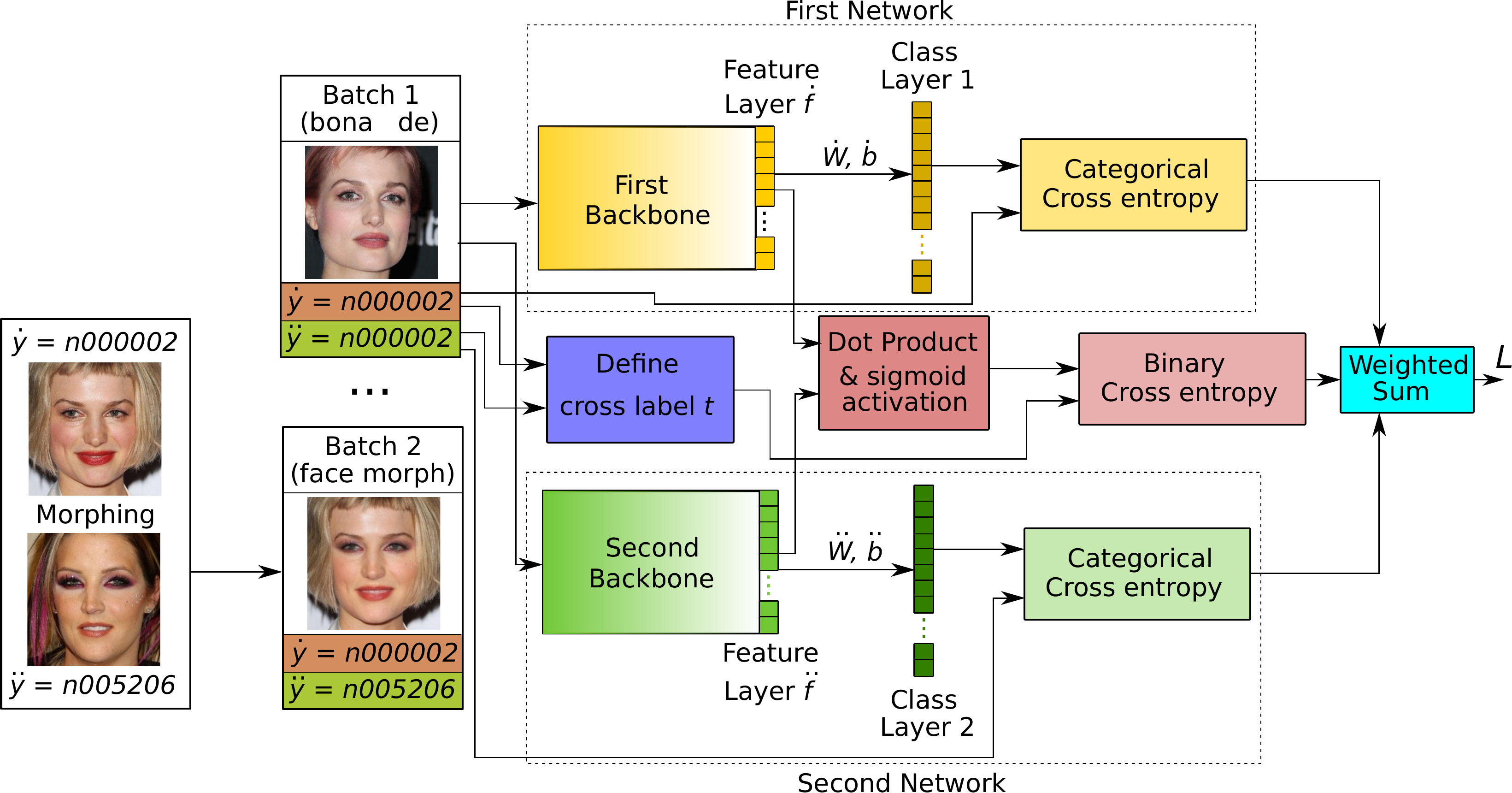}
\end{center}
   \caption{Schematic of the proposed method. For simplicity of visualisation batch contains a single image. Labels $\dot{y}$ and $\ddot{y}$ are indicated by names, when the real setup utilizes their numerical index value, which is encoded to one-hot vector.}
\label{fig:SSC_schematic}
\end{figure*}

In our work, we consider face morphing detection from the perspective of face recognition. In the case of following the approach via identity classification, face morphing introduces a problem, since a face morph image indeed belongs to several identities, which leads to the ambiguity of proper class labelling. 
In this work, we address this issue (Section ~\ref{data_morphing_separate}) in search of the method for single image morphing detection.

\section{Methodology} 
\label{Methodology}

In this section, we describe our technique for single image morphing detection via deep learning.  

In our research, we intuitively tried to invent a setup that will allow learning high-level deep features, that also carry some information about their authenticity. This resulted in the schematic that includes two backbone CNN based networks that are trained in a similar manner but biased in a way to discriminate morphed and bona fide images. Namely, our idea implies training two parallel networks which consider bona fide samples similarly and morphed samples differently (see Fig.~\ref{fig:SSC_schematic}). 

Both networks learn high-level features via classification tasks, which are different in terms of identity labelling of face morphs. \textit{First Network} labels them by the original identity from the first source image, the \textit{Second Network} - by the second original label. 

The extracted features are also explicitly compared by similarity metric (which is the dot product due to the softmax properties) and the result is classified according to the ground truth authenticity label of the image (bona fide/morph).  


The identity classification parts of the training scheme act as a regularisation that retains the facial discriminability of feature layers. That is why for identity classification we utilize a standard softmax, which allows easier convergence in comparison with its modifications (like ArcFace\cite{arcface_paper}).

Following the common formulation of softmax, our training process is regularized by the losses:

\begin{equation}
    L_{1} = -\frac{1}{N}\sum_{i}^{N} \log (\frac{e^{\dot{W}_{\dot{y}_i}^{T}\dot{f}_{i}+\dot{b}_{\dot{y}_i}}}{ \sum_{j}^{C} e^{\dot{f}_{\dot{y}_j}}})
\end{equation}

\begin{equation}
    L_{2} = -\frac{1}{N}\sum_{i}^{N} \log (\frac{e^{\ddot{W}_{\ddot{y}_i}^{T}\ddot{f}_{i}+\ddot{b}_{\ddot{y}_i}}}{ \sum_{j}^{C} e^{\ddot{f}_{\ddot{y}_j}}})
\end{equation}
where $\left \{  \dot{f}_i, \ddot{f}_i \right \}$ denote the deep features of the $i-th$ sample,  $\left \{  \dot{y}_i, \ddot{y}_i \right \}$ are the indexes of the class of the $i-th$ sample,  $\left \{  \dot{W}, \ddot{W} \right \}$ and $\left \{  \dot{b}, \ddot{b} \right \}$ are weights and biases of last fully connected layer (respectively for the $\left \{  \textit{First}, \textit{Second} \right \}$ networks). $N$ is the number of samples in a batch and $C$ is the total number of classes.

The target driver of the training process tries to explicitly discriminate morph/non-morph images. The dot product of backbones outputs indicates the morphing detection score. It is activated by sigmoid function, and the corresponding loss is defined as binary cross-entropy:

\begin{equation}
\begin{split}
    L_{3} = -\frac{1}{N} \sum_{i}^{N} t\log \frac{1}{1+e^{-D}}+ \\
    + (1-t)\log \left( 1-\frac{1}{1+e^{-D}} \right),
\end{split}
\end{equation}
where closs-label $t$ is computed by a comparison of input class labels: 
\begin{equation}
    t = abs(sgn(\dot{y}_i-\ddot{y}_i)),
\end{equation}
and $D$ is a dot product of high level features extracted by \textit{First} and \textit{Second} backbones:

\begin{equation}
    D = \dot{f} \cdot \ddot{f}
\end{equation}

The result loss for optimisation is combined as a weighted sum:

\begin{equation}
    L = \alpha_{1}L_{1}+\alpha_{2}L_{2}+\beta L_{3}
\end{equation}

By minimizing this loss in the fused classification setup, we learn the discriminative facial features that are explicitly regularized for morphing detection.

At the testing stage, the identity classification parts of the network are redundant and may be removed from the setup. The morphing detection decision is made by thresholding the scalar product of the backbones outputs.

Although our strategy is adapted for single image morphing detection indeed it is naturally also suitable for differential verification scenario. In this case, \textit{First} and \textit{Second} networks shall receive respectively two images (enrolled and life capture) instead of the same single image.

\section{Datasets}\label{datasets}
The proposed methodology requires the large labelled face dataset with an accompaniment of morphed images of identities from this dataset.

The academic community still doesn't have public ID document compliant datasets which are large enough for efficient training of modern deep networks (as an example, one of the largest  FRGC\_V2\cite{FRGC_V2_dataset} contains only $\sim$50k images and $\sim$500 identities). That is why our strategy for this work is to utilize the wild dataset which is filtered by the criteria of \textit{suitability for face morphing}. Conceptually this approach indeed is not novel and was recently utilized in face morphing research \cite{Right_Faces_to_Morph, PW-MAD}. In this work, we introduce a technique for semi-automatic wild dataset filtering for our method. 

As a source wild dataset we use VGGFace2\cite{VGGface2}($\sim$3M images, $\sim$9k classes, $\sim$360 samples per class, Licence - CC BY-SA 4.0) due to large average number of samples per identity (in comparison to other popular wild face datasets like CASIA-WebFace\cite{casia_webface}, MS-Celeb-1M\cite{ms_celeb_face, MS-Celeb-1M_CleanList}, Glint360K\cite{Partial_fc}, WebFace260M\cite{WebFace260M}) in order to have enough samples per identity after filtering.

\subsection{Wild Dataset Filtering}
\label{dataset_filtering}

Our dataset filtering strategy is based on a thresholding by quality metrics. Following Tremoço \etal \cite{QualFace} we used a set of quality scores for labeling the face images in the dataset: Blur \cite{variance_of_laplacian}, FaceQNet \cite{faceQnet}, BRISQUE \cite{brisque}, Face Illumination \cite{fiiqa} and Pose \cite{pose}. This set of scores allow to discriminate and select samples by their natural quality (Blur, BRISQUE), ID documents suitability (FaceQNet), face image acquisition parameters (Illumination, Pose).

\begin{figure}[t] 

\begin{center}
  \includegraphics[width=0.99\linewidth]{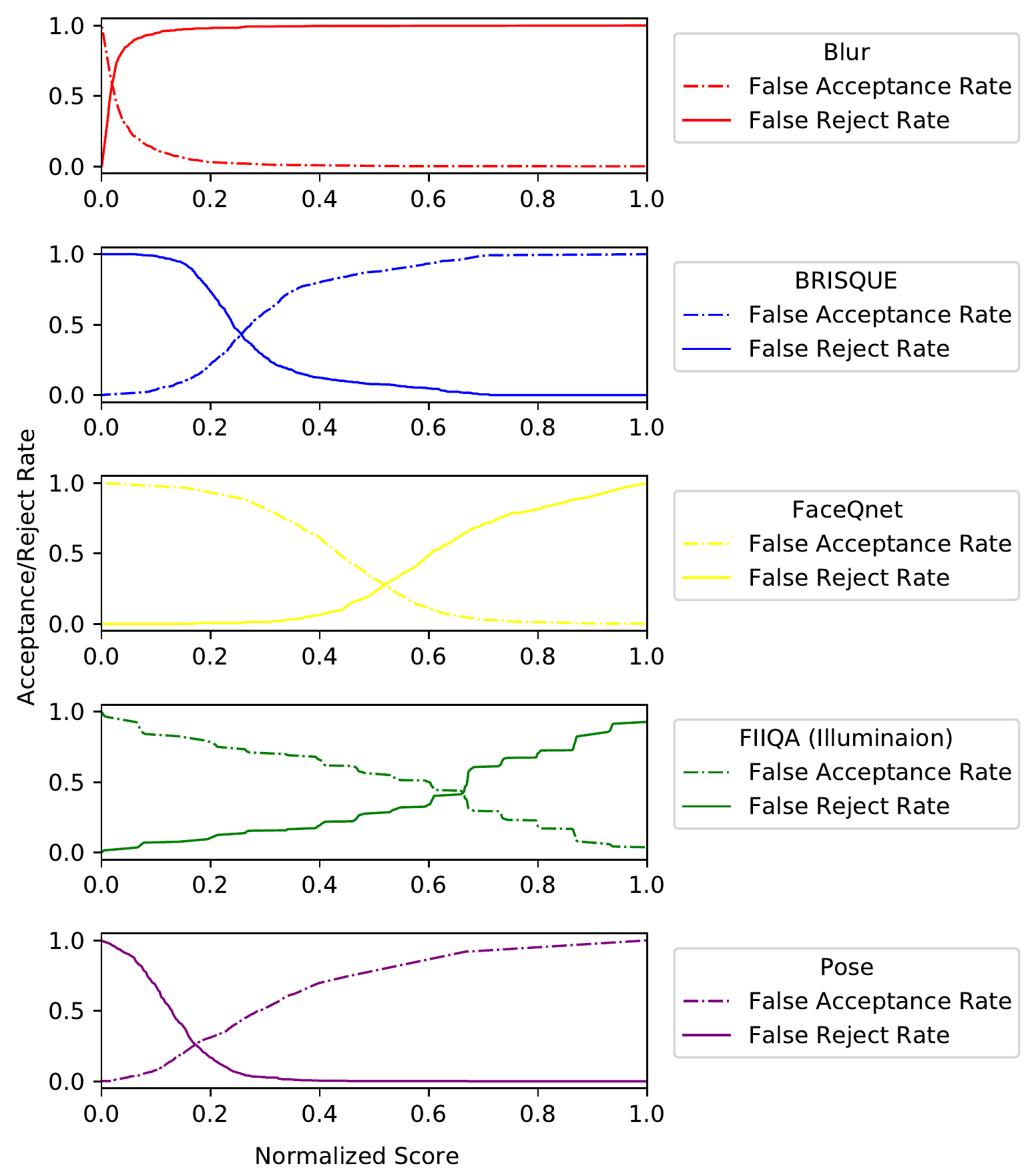}
\end{center}
  \caption{FAR and FRR for result manual quality labeling of random samples from VGGFace2.}
\label{fig:combined_scores}
\end{figure}

Next, we randomly select samples and manually label them with a binary value (accept/reject). This acceptance is defined by the criteria of suitability for application in face morphing (namely by user's choice). In our setup, we assure that samples are selected distributively across the quality scores values. Namely, we split the total quality score range into a set of sub-ranges and define the minimum number of samples to be selected from each sub-range. By proceeding around 4k images in our setup, we harvest the dependency of FAR (False Acceptance Rate) and FRR (False Rejection Rate) from quality scores values (see Fig.~\ref{fig:combined_scores}).

\begin{figure}[htbp] 
\begin{center}
  \includegraphics[width=0.99\linewidth]{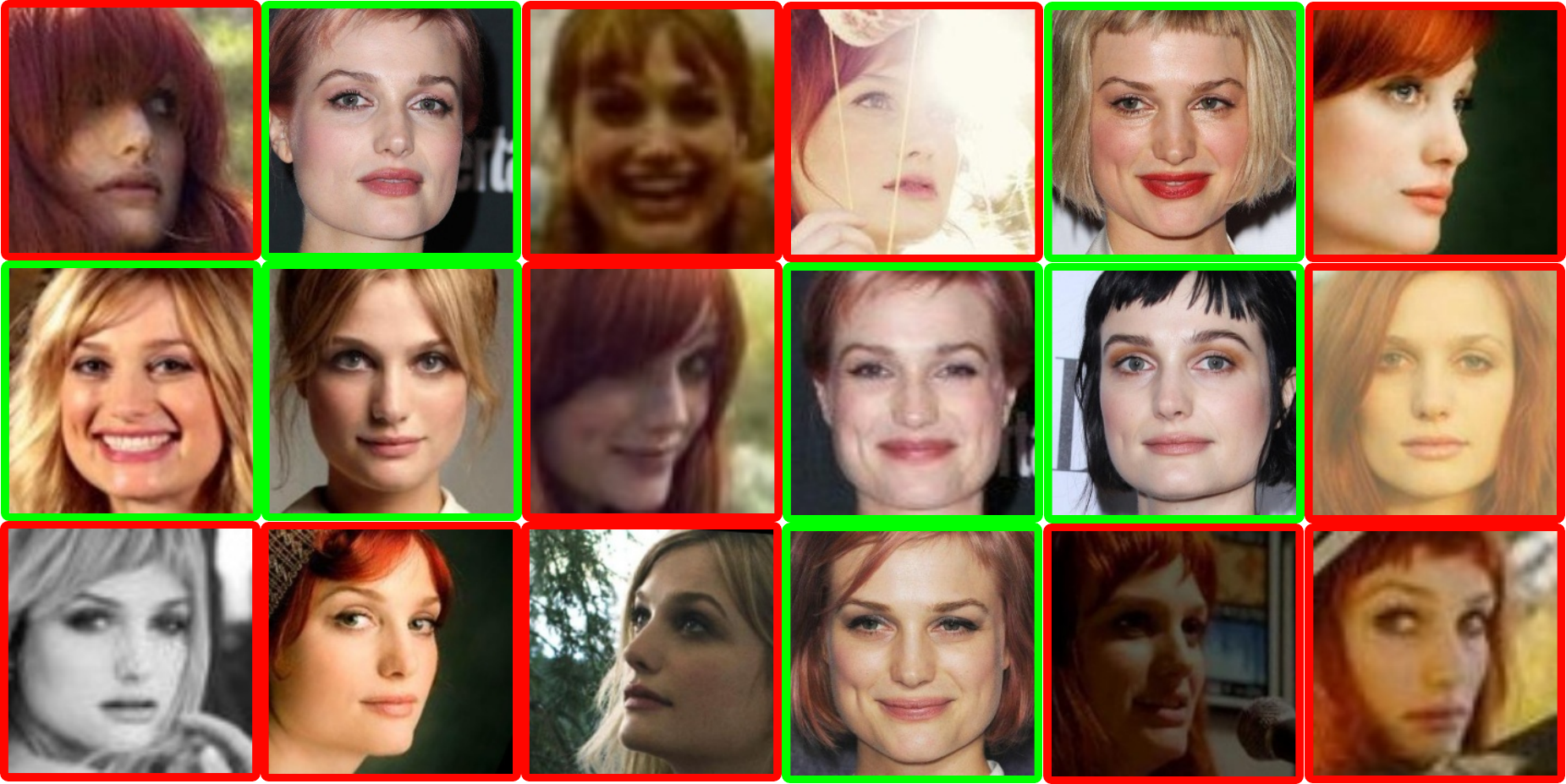}
\end{center}
  \caption{Example of VGGFace2 filtering result. Accepted images (green box) and Rejected images (red box).}
\label{fig:wild_filtering_result}
\end{figure}


The dataset filtering is then performed with joint thresholding by those quality metrics. For each score, we select the threshold at a point of EER (Equal Error Rate).
As a result we get the \textit{VGGFace2-selected} dataset with the same identity list as original source and around 500k images (see Fig.~\ref{fig:wild_filtering_result}).

\subsection{Morph Dataset Harvesting Strategy}
\label{data_morphing_separate}
\begin{figure}[htbp] 
\begin{center}
  \includegraphics[width=0.7\linewidth]{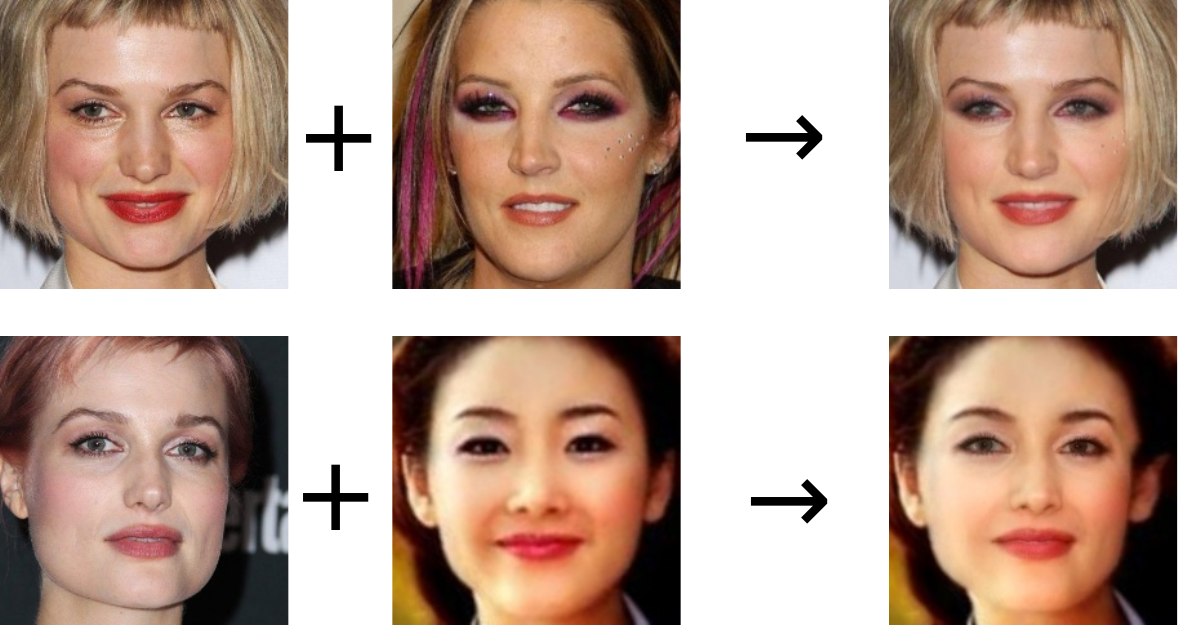}
\end{center}
  \caption{Examples of generated morphs with landmark based approach. Background is restored by one of the source images (chosen randomly).}
\label{fig:morph_result}
\end{figure}

For application in our method, the filtered wild dataset is needed to be accompanied by a large collection of face morphs. 
We automatically generate these images with our customized landmark-based morphing approach (with blending coefficient $\alpha = 0.5$) (see Fig.~\ref{fig:morph_result}).

A key requirement for effective learning is to provide unambiguity of proper class labelling in our training method.
Namely after generating face morph from two arbitrary samples of the original dataset, the resulting image indeed belongs to both source identities. That is why fully random image pairing (for generating morphs) will result in classification confusion.

To avoid that we utilize the following strategy. First, we separate the total list of identities into two disjoint parts, which are attributed to the \textit{First} and the \textit{Second} networks respectively. Next, to generate a face morph, we randomly pair images from identities of these list halves. Each generated image is then labelled according to the attributed sublist for classification by the \textit{First} and the \textit{Second} networks.
Let us note that this labelling is made differently for each morphed image and similarly for bona fide images in both networks (see Fig.~\ref{fig:SSC_schematic}). That is why this technique, which primarily acts as a regularisation, also amplifies the morphing detection performance.

Following the above procedure, we generate \textit{VGGFace2-selected-morph} dataset, which contains around 1M morphed images.

\subsection{Selfmorphing}
The fully automatic landmark morphing methods may introduce a number of visible artefacts to the generated images (like blending artefacts). That is why without additional regularisation our method will be biased to learning those artefacts, which is not a realistic scenario. Real fraudulent morphs are retouched with the intention to remove any perceptual artefacts. 

To address this problem, we utilize \textit{selfmorphs}, which are generated by applying face morphing to images of the same identity. This concept was indeed recently introduced by Borghi \etal
\cite{morphing_artifacts_retouching} and was used for generating images with visible artefacts. Then for removing these artefacts the authors trained the Conditional GAN using original images as a ground truth reference. 

In this work, we utilize selfmorph images to focus morphing detection onto the deep face features behaviour, rather than to detecting artefacts. We assume that deep discriminative face features remain after performing selfmorphing.  In the proposed method schematic (Fig.~\ref{fig:SSC_schematic}) we consider selfmorphs as bona fide samples. 

\begin{figure}[htbp] 
\begin{center}
  \includegraphics[width=0.99\linewidth]{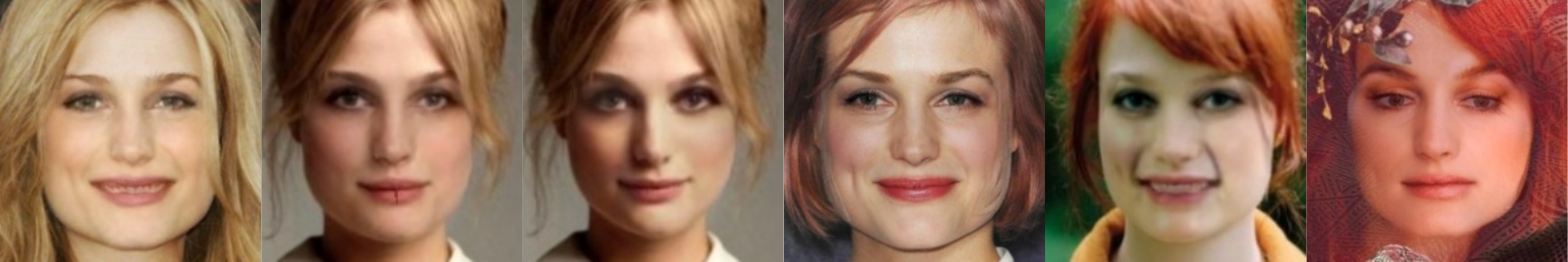}
\end{center}
  \caption{Examples of generated selfmorphs. Images contain blending artefacts but the identity is perceptually retained. }
\label{fig:self_morph_result}
\end{figure}

We perform a random pairing of samples within each identity from the \textit{VGGFace2-selected} and generate \textit{VGGFace2-selected-selfmorph} dataset, which contains around 500k images (see Fig.~\ref{fig:self_morph_result}).

\begin{figure*}
\begin{center}
  \includegraphics[width=0.99\linewidth]{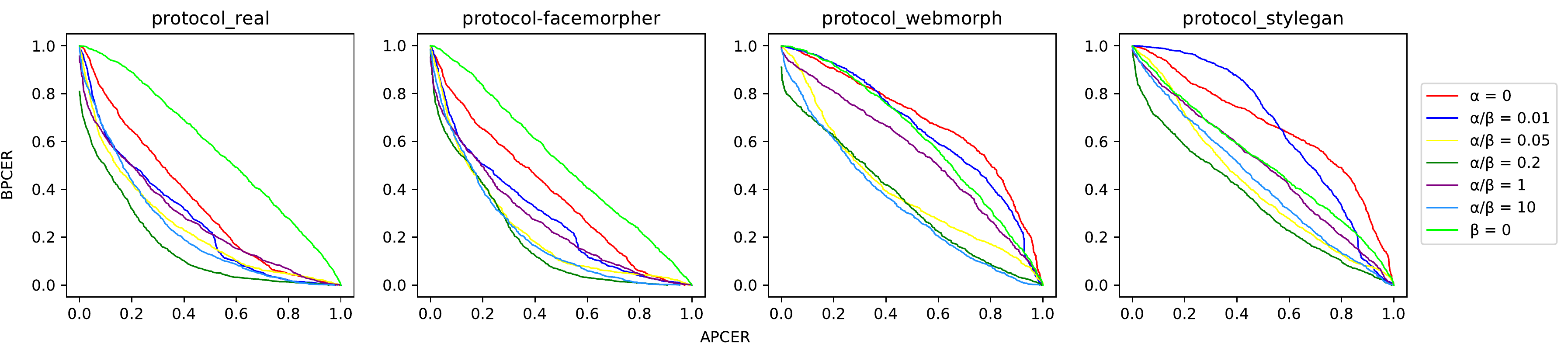}
\end{center}
\vspace*{-8pt}
\caption{Detection Error Trade-off curves for various $\alpha/\beta$ proportions in different protocols.}
\label{fig:combined_lin}
\end{figure*}

\begin{figure*}
\vspace*{-8pt}
\begin{center}
  \includegraphics[width=0.99\linewidth]{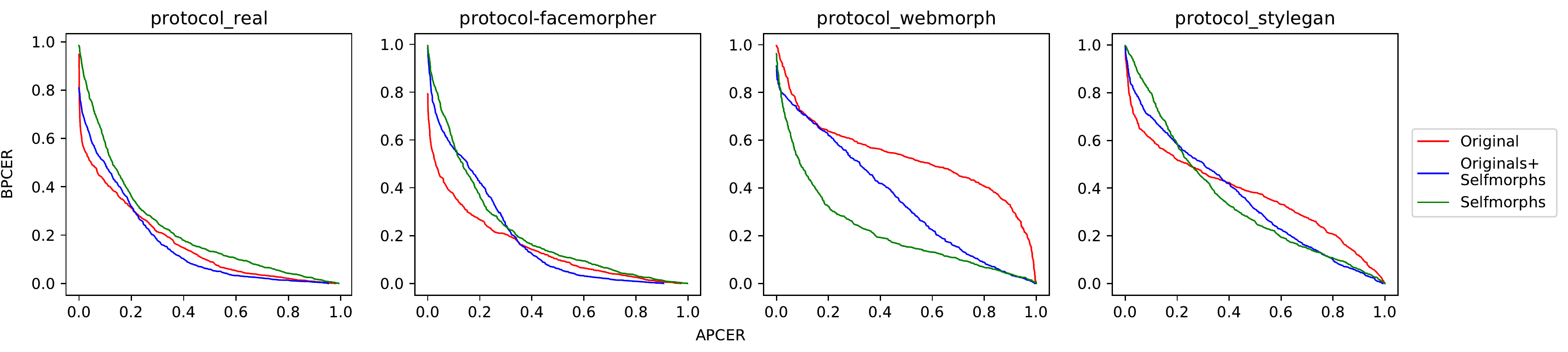}
\end{center}
\vspace*{-8pt}
\caption{Detection Error Trade-off curves for various bona fide images selection in different protocols.}
\label{fig:self_test}
\end{figure*}

\section{Benchmarking} \label{m_benchmark}

There are few public benchmarks for evaluating the performance of morphing detection or morphing resistant algorithms: the NIST FRVT MORPH  \cite{bench_NIST_morph} and  FVC-onGoing MAD\cite{bench_1,bench_1_2}.     
Both of these benchmarks accept no-reference and differential morphing algorithms, however, they are proprietary and are executed on the maintainer side. Thus they have a number of submission restrictions.  

The straightforward metric for evaluating single image morphing detection is the dependency of Bona fide Presentation Classification Error Rate (BPCER) from Attack Presentation Classification Error Rate (APCER) (according to ISO/IEC 30107-3 \cite{REFERENCE_OF_ISO_30107}), which may be plotted as a Detection Error Trade-off (DET) curve.

\subsection{Face Morphing Detection Benchmark}
For this work, we  develop an custom benchmark, which is to be executed on the developer side. \textit{We are not making it public at this stage}). 
The existing public benchmarks provide useful data but usually specify the protocols for the certain software frameworks \cite{FRLL_FRGC_Morphs}.

Our benchmark intends to provide the functionality for estimating the morphing detection performance, for generating custom protocols and also for further comparison of the results from different developers with existing protocols.
At this stage of our work, we focus on the single image morphing detection with only the usage of public data (however, we assume the possibility of further adapting private datasets).   

We generate several protocols for single image morphing detection. Our benchmark is based on the FRGC-Morphs, FRLL-Morphs \cite{FRLL_FRGC_Morphs}, AMSL \cite{amsl_morph} and Dustone datasets\cite{dustone_dataset}.
Using this data we combine several benchmark protocols with various types of face morphs:

\begin{itemize}[noitemsep,topsep=1pt]
	\item \textit{protocol-real} ($\sim$3k morphs(Dustone+AMSL)), which includes  morphs with low level of visible blending artifacts, and imitates realistic presentation attacks.
	\item \textit{protocol-facemorpher} ($\sim$2k morphs), which includes simple morphs with foreground and background artifacts
	\item \textit{protocol-webmorph} ($\sim$1k morphs), which includes images with background artifacts but the low level of artifacts inside the face contour
	\item \textit{protocol-stylegan} ($\sim$2k morphs), which includes StyleGan morphs
\end{itemize}

As bona fide images all our protocols use frontal faces from the following public datasets: FRLL Set\cite{Face_Research_Lab_London_Set}, FEI \cite{FEI_dataset}, AR\cite{AR_face}, Aberdeen and Utrecht \cite{Pics_dataset} ($\sim$1.5k images in total).

\section{Experiments and Results}
To analyze the performance of our approach we perform several experiments with our method.
As backbone networks we use ResNet-50 \cite{Resnet2016}, which are initialized with weights, pretrained on the ImageNet dataset. Followed by pooling and dropout layers each backbone returns 512 deep features.  Input images (RGB 3-channel) are aligned and resized to 224$\times$224.
We report the performance by $APCER@BPCER=(0.1, 0.01)$ and $BPCER@APCER=(0.1, 0.01)$.

\begin{table*}[t]
\begin{center}
\begin{tabular}{|c|cccccccc|}
\hline
\multirow{3}{*}{Method} & \multicolumn{8}{c|}{$APCER@BPCER=\delta$}                                                                                                                                                                                   \\ \cline{2-9} 
                        & \multicolumn{2}{c|}{protocol-real}                                  & \multicolumn{2}{c|}{protocol-facemorpher}                                  & \multicolumn{2}{c|}{protocol-webmorph}                                  & \multicolumn{2}{c|}{protocol-stylegan}             \\ \cline{2-9} 
                        & \multicolumn{1}{c|}{$\delta = 0.1$} & \multicolumn{1}{c|}{$\delta = 0.01$} & \multicolumn{1}{c|}{$\delta = 0.1$} & \multicolumn{1}{c|}{$\delta = 0.01$} & \multicolumn{1}{c|}{$\delta = 0.1$} & \multicolumn{1}{c|}{$\delta = 0.01$} & \multicolumn{1}{c|}{$\delta = 0.1$} & {$\delta = 0.01$} \\ \hline
$\alpha = 0$                       & \multicolumn{1}{c|}{0.697}      & \multicolumn{1}{c|}{0.947}       & \multicolumn{1}{c|}{0.756}      & \multicolumn{1}{c|}{0.939}       & \multicolumn{1}{c|}{0.976}      & \multicolumn{1}{c|}{0.995}       & \multicolumn{1}{c|}{0.980}      &    0.998    \\ \hline
$\alpha/\beta = 0.01$                          & \multicolumn{1}{c|}{0.601}      & \multicolumn{1}{c|}{0.895}       & \multicolumn{1}{c|}{0.651}      & \multicolumn{1}{c|}{0.957}       & \multicolumn{1}{c|}{0.945}      & \multicolumn{1}{c|}{0.992}       & \multicolumn{1}{c|}{0.895}      &    0.991    \\ \hline
$\alpha/\beta = 0.05$                       & \multicolumn{1}{c|}{0.607}      & \multicolumn{1}{c|}{0.965}       & \multicolumn{1}{c|}{0.502}      & \multicolumn{1}{c|}{0.968}       & \multicolumn{1}{c|}{0.915}      & \multicolumn{1}{c|}{0.999}       & \multicolumn{1}{c|}{0.842}      &    0.996    \\ \hline
$\alpha/\beta = 0.2$                & \multicolumn{1}{c|}{0.401}      & \multicolumn{1}{c|}{0.835}       & \multicolumn{1}{c|}{0.431}      & \multicolumn{1}{c|}{0.786}       & \multicolumn{1}{c|}{0.778}      & \multicolumn{1}{c|}{0.979}       & \multicolumn{1}{c|}{0.799}      &    0.969    \\ \hline
$\alpha/\beta = 1$                    & \multicolumn{1}{c|}{0.711}      & \multicolumn{1}{c|}{0.935}       & \multicolumn{1}{c|}{0.670}      & \multicolumn{1}{c|}{0.928}       & \multicolumn{1}{c|}{0.942}      & \multicolumn{1}{c|}{0.995}       & \multicolumn{1}{c|}{0.913}      &     0.982   \\ \hline
$\alpha/\beta = 10$                     & \multicolumn{1}{c|}{0.556}      & \multicolumn{1}{c|}{0.839}       & \multicolumn{1}{c|}{0.513}      & \multicolumn{1}{c|}{0.791}       & \multicolumn{1}{c|}{0.749}      & \multicolumn{1}{c|}{0.923}       & \multicolumn{1}{c|}{0.852}      &    0.969    \\ \hline
$\beta = 0.0$                     & \multicolumn{1}{c|}{0.945}      & \multicolumn{1}{c|}{0.996}       & \multicolumn{1}{c|}{0.922}      & \multicolumn{1}{c|}{0.993}       & \multicolumn{1}{c|}{0.954}      & \multicolumn{1}{c|}{0.997}       & \multicolumn{1}{c|}{0.949}      &     0.997   \\ \hline \hline
$\alpha/\beta = 0.2$ Original                 & \multicolumn{1}{c|}{0.481}      & \multicolumn{1}{c|}{0.875}       & \multicolumn{1}{c|}{0.498}      & \multicolumn{1}{c|}{0.876}       & \multicolumn{1}{c|}{0.987}      & \multicolumn{1}{c|}{0.997}       & \multicolumn{1}{c|}{0.915}      &    0.993    \\ \hline
$\alpha/\beta = 0.2$ Selfmorphs                   & \multicolumn{1}{c|}{0.608}      & \multicolumn{1}{c|}{0.938}       & \multicolumn{1}{c|}{0.568}      & \multicolumn{1}{c|}{0.911}       & \multicolumn{1}{c|}{0.704}      & \multicolumn{1}{c|}{0.986}       & \multicolumn{1}{c|}{0.816}      &    0.983    \\ \hline
\end{tabular}
\end{center}
\vspace*{-3pt}
\caption{APCER@BPCER = (0.1, 0.01) of our method for various $\alpha/\beta$ proportions and bona fide images selection in different protocols.}
\label{tab:APCER_BPCER_results}
\end{table*}

\begin{table*}[t]
\vspace*{-3pt}
\begin{center}
\begin{tabular}{|c|cccccccc|}
\hline
\multirow{3}{*}{Method} & \multicolumn{8}{c|}{$BPCER@APCER=\delta$}                                                                                                                                                                                   \\ \cline{2-9} 
                        & \multicolumn{2}{c|}{protocol-real}                                  & \multicolumn{2}{c|}{protocol-facemorpher}                                  & \multicolumn{2}{c|}{protocol-webmorph}                                  & \multicolumn{2}{c|}{protocol-stylegan}             \\ \cline{2-9} 
                        & \multicolumn{1}{c|}{$\delta = 0.1$} & \multicolumn{1}{c|}{$\delta = 0.01$} & \multicolumn{1}{c|}{$\delta = 0.1$} & \multicolumn{1}{c|}{$\delta = 0.01$} & \multicolumn{1}{c|}{$\delta = 0.1$} & \multicolumn{1}{c|}{$\delta = 0.01$} & \multicolumn{1}{c|}{$\delta = 0.1$} & {$\delta = 0.01$} \\ \hline
$\alpha = 0$                       & \multicolumn{1}{c|}{0.795}      & \multicolumn{1}{c|}{0.993}       & \multicolumn{1}{c|}{0.781}      & \multicolumn{1}{c|}{0.971}       & \multicolumn{1}{c|}{0.961}      & \multicolumn{1}{c|}{0.998}       & \multicolumn{1}{c|}{0.952}      &     0.998   \\ \hline
$\alpha/\beta = 0.01$                          & \multicolumn{1}{c|}{0.625}      & \multicolumn{1}{c|}{0.968}       & \multicolumn{1}{c|}{0.638}      & \multicolumn{1}{c|}{0.979}       & \multicolumn{1}{c|}{0.969}      & \multicolumn{1}{c|}{0.997}       & \multicolumn{1}{c|}{0.991}      &     1.0   \\ \hline
$\alpha/\beta = 0.05$                       & \multicolumn{1}{c|}{0.577}      & \multicolumn{1}{c|}{0.950}       & \multicolumn{1}{c|}{0.598}      & \multicolumn{1}{c|}{0.951}       & \multicolumn{1}{c|}{0.841}      & \multicolumn{1}{c|}{0.989}       & \multicolumn{1}{c|}{0.895}      &     0.991   \\ \hline
$\alpha/\beta = 0.2$                     & \multicolumn{1}{c|}{0.494}      & \multicolumn{1}{c|}{0.726}       & \multicolumn{1}{c|}{0.562}      & \multicolumn{1}{c|}{0.848}       & \multicolumn{1}{c|}{0.710}      & \multicolumn{1}{c|}{0.822}       & \multicolumn{1}{c|}{0.697}      &      0.904  \\ \hline
$\alpha/\beta = 1$                    & \multicolumn{1}{c|}{0.616}      & \multicolumn{1}{c|}{0.871}       & \multicolumn{1}{c|}{0.628}      & \multicolumn{1}{c|}{0.843}       & \multicolumn{1}{c|}{0.882}      & \multicolumn{1}{c|}{0.963}       & \multicolumn{1}{c|}{0.851}      &     0.960   \\ \hline
$\alpha/\beta = 10$                     & \multicolumn{1}{c|}{0.642}      & \multicolumn{1}{c|}{0.892}       & \multicolumn{1}{c|}{0.604}      & \multicolumn{1}{c|}{0.913}       & \multicolumn{1}{c|}{0.742}      & \multicolumn{1}{c|}{0.931}       & \multicolumn{1}{c|}{0.829}      &    0.967    \\ \hline
$\beta = 0.0$                     & \multicolumn{1}{c|}{0.956}      & \multicolumn{1}{c|}{0.998}       & \multicolumn{1}{c|}{0.932}      & \multicolumn{1}{c|}{0.998}       & \multicolumn{1}{c|}{0.971}      & \multicolumn{1}{c|}{0.998}       & \multicolumn{1}{c|}{0.874}      &    0.986    \\ \hline \hline
$\alpha/\beta = 0.2$ Original                 & \multicolumn{1}{c|}{0.417}      & \multicolumn{1}{c|}{0.601}       & \multicolumn{1}{c|}{0.370}      & \multicolumn{1}{c|}{0.595}       & \multicolumn{1}{c|}{0.718}      & \multicolumn{1}{c|}{0.963}       & \multicolumn{1}{c|}{0.605}      &    0.862   \\ \hline
$\alpha/\beta = 0.2$ Selfmorphs                   & \multicolumn{1}{c|}{0.580}      & \multicolumn{1}{c|}{0.912}       & \multicolumn{1}{c|}{0.587}      & \multicolumn{1}{c|}{0.905}       & \multicolumn{1}{c|}{0.484}      & \multicolumn{1}{c|}{0.842}       & \multicolumn{1}{c|}{0.798}      &    0.976    \\ \hline
\end{tabular}
\end{center}
\vspace*{-3pt}
\caption{BPCER@APCER = (0.1, 0.01) of our method for various $\alpha/\beta$ proportions and bona fide images selection in different protocols.}
\label{tab:BPCER_APCER_results}
\end{table*}

Our default training dataset is a joined and shuffled concatenation of \textit{VGGFace2-selected}, \textit{VGGFace2-selected-selfmorph}, and \textit{VGGFace2-selected-morph}.
It is important to note that in all experiments we assured the equal balance between the numbers of morphed and non-morphed (which are bona fide + selfmorphed) images in the training dataset.

\subsection{Fused Classification Balance}
\label{fused_class_balance}
For effective convergence and further morphing detection, our method requires choosing the proper balance between the elements of the loss function. Namely the balance between $\alpha$ $(=\alpha_1=\alpha_2)$ and $\beta$ (disbalance of $\alpha_1$ and $\alpha_2$ didn't demonstrate any interesting behaviour in our tests).
We perform training of our method with different proportional settings also including the ablation of particular parts from the overall loss. 
Our experiments demonstrate (see Fig. \ref{fig:combined_lin} and Tab. ~\ref{tab:APCER_BPCER_results}, \ref{tab:BPCER_APCER_results}) that by varying $\alpha / \beta$ proportion it is possible to achieve some optimal performance of morphing detection in different protocols. Our strategy allows generalizing the detection of morphing even to the images, which are generated with GANs even accounting that this type of morphing is totally unseen in the training. 



On the edge case with excluded main loss function driver (namely binary morph/bona fide classification), our method demonstrates the almost random detection decision.
At the same time, ablation of the regularisation ($\beta = 0$) also leads to bad performance, which we relate with the overfitting on the trivial binary classification learning task.

Summing up, we can conclude that our strategy allows learning such face features which are discriminative by the criteria of authenticity.

\subsection{Data combination experiments}
Further experiments are performed with the selected proportion $\alpha / \beta = 0.2$ in order to understand the impact of selfmorphing for our method. 

In comparison to the dataset selection in Section \ref{fused_class_balance} where the collection of bona fide samples is split evenly to original and selfmorphs, we test two more options where these particular parts are ablated from the total dataset.

\begin{table*}[t]
\begin{center}
\begin{tabular}{|c|cccccccc|}
\hline
\multirow{3}{*}{Method} & \multicolumn{8}{c|}{$APCER@BPCER=\delta$}                                                                                                                                                                                   \\ \cline{2-9} 
                        & \multicolumn{2}{c|}{P1}                                  & \multicolumn{2}{c|}{P2}                                  & \multicolumn{2}{c|}{P3}                                  & \multicolumn{2}{c|}{P4}             \\ \cline{2-9} 
                        & \multicolumn{1}{c|}{$\delta = 0.1$} & \multicolumn{1}{c|}{$\delta = 0.01$} & \multicolumn{1}{c|}{$\delta = 0.1$} & \multicolumn{1}{c|}{$\delta = 0.01$} & \multicolumn{1}{c|}{$\delta = 0.1$} & \multicolumn{1}{c|}{$\delta = 0.01$} & \multicolumn{1}{c|}{$\delta = 0.1$} & {$\delta = 0.01$} \\ \hline

Aghdaie \etal \cite{AttentionMorphing}                      & \multicolumn{1}{c|}{0.965}      & \multicolumn{1}{c|}{0.998}       & \multicolumn{1}{c|}{0.923}      & \multicolumn{1}{c|}{0.991}       & \multicolumn{1}{c|}{\textbf{0.015}}      & \multicolumn{1}{c|}{\textbf{0.200}}       & \multicolumn{1}{c|}{\textbf{0.271}}      &   \textbf{0.721}   \\ \hline
Debiasi \etal \cite{PRNU_2} & \multicolumn{1}{c|}{\textbf{0.049} }      & \multicolumn{1}{c|}{0.823}       & \multicolumn{1}{c|}{0.994}      & \multicolumn{1}{c|}{1.000}       & \multicolumn{1}{c|}{1.000}      & \multicolumn{1}{c|}{1.000}       & \multicolumn{1}{c|}{0.985}      &     0.994   \\ \hline
Ramachandra  \etal \cite{Towards_morphing_detection}                       & \multicolumn{1}{c|}{0.375}      & \multicolumn{1}{c|}{0.990}       & \multicolumn{1}{c|}{0.938}      & \multicolumn{1}{c|}{0.985}       & \multicolumn{1}{c|}{0.159}      & \multicolumn{1}{c|}{0.998}       & \multicolumn{1}{c|}{0.936 }      &      0.996  \\ \hline
Scherhag \etal \cite{Multi_Algorithm_FUsion_Morphing}                    & \multicolumn{1}{c|}{1.000}      & \multicolumn{1}{c|}{1.000}       & \multicolumn{1}{c|}{0.997 }      & \multicolumn{1}{c|}{1.000}       & \multicolumn{1}{c|}{0.996}      & \multicolumn{1}{c|}{1.000}       & \multicolumn{1}{c|}{0.993}      &    1.000    \\ \hline
Lorenz  \etal \cite{morphing_fusion}               & \multicolumn{1}{c|}{0.380}      & \multicolumn{1}{c|}{1.000}       & \multicolumn{1}{c|}{0.966}      & \multicolumn{1}{c|}{1.000}       & \multicolumn{1}{c|}{0.819}      & \multicolumn{1}{c|}{1.000}       & \multicolumn{1}{c|}{0.971 }      &    0.995    \\ \hline
Ferrara \etal \cite{unibo}                   & \multicolumn{1}{c|}{0.477}      & \multicolumn{1}{c|}{0.999}       & \multicolumn{1}{c|}{0.978}      & \multicolumn{1}{c|}{1.000}       & \multicolumn{1}{c|}{0.037}      & \multicolumn{1}{c|}{0.810}       & \multicolumn{1}{c|}{0.420 }      &     0.777   \\ \hline
Ours                    & \multicolumn{1}{c|}{0.434 }      & \multicolumn{1}{c|}{\textbf{0.686}}       & \multicolumn{1}{c|}{\textbf{0.842}}      & \multicolumn{1}{c|}{\textbf{0.954}}       & \multicolumn{1}{c|}{0.323}      & \multicolumn{1}{c|}{0.639}       & \multicolumn{1}{c|}{0.499}      &    0.805   \\ \hline
\end{tabular}
\end{center}
\vspace*{-3pt}
\caption{Comparison with the state of the art single image morphing detection methods by APCER@BPCER = (0.1, 0.01).}
\label{tab:sota_comparison_single}
\end{table*}

\begin{table*}[t]
\vspace*{-3pt}
\begin{center}
\begin{tabular}{|c|cccccccc|}
\hline
\multirow{3}{*}{Method} & \multicolumn{8}{c|}{$APCER@BPCER=\delta$}                                                                                                                                                                                   \\ \cline{2-9} 
                        & \multicolumn{2}{c|}{\textit{P1}}                                  & \multicolumn{2}{c|}{\textit{P2}}                                  & \multicolumn{2}{c|}{\textit{P3}}                                  & \multicolumn{2}{c|}{\textit{P4}}             \\ \cline{2-9} 
                        & \multicolumn{1}{c|}{$\delta = 0.1$} & \multicolumn{1}{c|}{$\delta = 0.01$} & \multicolumn{1}{c|}{$\delta = 0.1$} & \multicolumn{1}{c|}{$\delta = 0.01$} & \multicolumn{1}{c|}{$\delta = 0.1$} & \multicolumn{1}{c|}{$\delta = 0.01$} & \multicolumn{1}{c|}{$\delta = 0.1$} & {$\delta = 0.01$} \\ \hline
Scherhag \etal \cite{face_morphing_MAD}                         & \multicolumn{1}{c|}{\textbf{0.109}}      & \multicolumn{1}{c|}{1.000}       & \multicolumn{1}{c|}{\textbf{0.094}}      & \multicolumn{1}{c|}{1.000}       & \multicolumn{1}{c|}{\textbf{0.031}}      & \multicolumn{1}{c|}{1.000}       & \multicolumn{1}{c|}{\textbf{0.093 }}      &   1.000     \\ \hline
Lorenz  \etal \cite{morphing_fusion}                        & \multicolumn{1}{c|}{0.432 }      & \multicolumn{1}{c|}{1.000}       & \multicolumn{1}{c|}{0.634 }      & \multicolumn{1}{c|}{1.000}       & \multicolumn{1}{c|}{0.168}      & \multicolumn{1}{c|}{1.000}       & \multicolumn{1}{c|}{0.732 }      &   1.000     \\ \hline
Scherhag \etal \cite{Multi_Algorithm_FUsion_Morphing}                        & \multicolumn{1}{c|}{0.208}      & \multicolumn{1}{c|}{1.000}       & \multicolumn{1}{c|}{0.927}      & \multicolumn{1}{c|}{1.000}       & \multicolumn{1}{c|}{0.451}      & \multicolumn{1}{c|}{1.000}       & \multicolumn{1}{c|}{0.934}      &    1.000    \\ \hline
Ours                    & \multicolumn{1}{c|}{0.865 }      & \multicolumn{1}{c|}{\textbf{0.967}}       & \multicolumn{1}{c|}{0.948}      & \multicolumn{1}{c|}{\textbf{0.981}}       & \multicolumn{1}{c|}{0.815}      & \multicolumn{1}{c|}{\textbf{0.929}}       & \multicolumn{1}{c|}{0.861}      &   \textbf{0.987}     \\ \hline
\end{tabular}
\end{center}
\vspace*{-3pt}
\caption{Comparison with the state of the art differential morphing detection methods by APCER@BPCER = (0.1, 0.01).}
\label{tab:sota_comparison_differential}
\end{table*}

Our results (see Fig. \ref{fig:self_test} and Tab. \ref{tab:APCER_BPCER_results},\ref{tab:BPCER_APCER_results}) proves the significant importance of selfmorphs in our strategy. Utilizing selfmorphs at the training stage allows to reduce the emphasis of the detection of facial blending artefacts and shift it to the behaviour of the deep feature for generalizing to unseen types of attacks.

\subsection{NIST FRVT MORPH Results}
We have performed the comparison of the results of our method and the state of the art face morphing detection approaches with NIST FRVT MORPH Benchmark (Report of October 28, 2021)\cite{bench_NIST_morph}. 

We select the model with $\alpha / \beta = 0.2$ from Section \ref{fused_class_balance} as our best model and present results of comparison in several protocols:
\begin{itemize}[noitemsep,topsep=1pt]
	\item \textit{P1} - Visa-Border (25727 Morphs)
	\item \textit{P2} - Manual (323 Morphs)
	\item \textit{P3} - MIPGAN-II (2464 Morphs)
	\item \textit{P4} - Print + Scanned (3604 Morphs)
\end{itemize}

As bona fide samples all protocols utilize a large collection of 1047389 Bona Fide images.
The comparison is performed by the metrics $APCER@BPCER = (0.1, 0.01)$.

\subsection{Single Image MAD}

First, we perform the comparison in the target single image morphing detection scenario (see Table ~\ref{tab:sota_comparison_single}).

MorDeephy outperforms other techniques in detecting landmark-based morphs and challenging manual morphs and achieves comparable results in other protocols. 

Also, our method does not demonstrate bias to a particular morphing generative strategy and has the most stable performance across all protocols in comparison to other approaches.

It is important to note that these results are achieved by utilizing a rather straightforward and simple morphing technique during training (without any adaptation to realistic scenario or modifications for removing artefacts), which proves that our method allows generalizing morphing detection to various unseen generative approaches by focusing on deep face features behaviour.

\subsection{Differential MAD}
The suitability of the approach for the differential scenario was previously mentioned. We perform the straightforward application of our method to the differential detection and compare with several SOTA methods (see Table ~\ref{tab:sota_comparison_differential}). In order to do it the \textit{Second network} (see Fig.~\ref{fig:SSC_schematic}) receives the life capture image (instead of the same image as the \textit{First network}) on the testing stage.

The comparison demonstrates that our method has more regular characteristics and outperforms other ones by APCER on low demanding BPCER.
These results are achieved with zero effort of training our method in a differential manner, which proves that the extracted deep authentically discriminative features are not only characteristic for a particular sample but are generalised to the identity.

\section{Conclusion}

We introduce a  novel deep learning strategy for single image face morphing detection,  which implies utilising a complex classification task. It is directed onto learning the deep facial features, which carry information about the authenticity of these features.
Our method achieved the state of the art performance and demonstrated a prominent ability for generalising the task of morphing detection to unseen scenarios (like GAN morphs and print/scan morphs).

Our work also introduces several additional contributions, which are the public and easy-to-use face morphing detection benchmark and the results of our wild datasets filtering strategy.

In our further work, we will focus on improving the performance by applying more sophisticated morphing techniques during training and on explicit adapting our method to the differential scenario, which will require sophisticated sampling strategies.


{\small
\bibliographystyle{ieee_fullname}
\bibliography{MorDeephy}
}

\end{document}